# Automated design of collective variables using supervised machine learning

Mohammad M. Sultan[1], & Vijay S. Pande[2†]

[1]Department of Chemistry, Stanford University, 318 Campus Drive, Stanford, California 94305, USA.

[2]Department of Bioengineering, Stanford University, 318 Campus Drive, Stanford, California 94305, USA.

[†]Pande@stanford.edu

**Abstract:**
Selection of appropriate collective variables for enhancing sampling of molecular simulations remains an unsolved problem in computational modeling. In particular, picking initial collective variables (CVs) is particularly challenging in higher dimensions. Which atomic coordinates or transforms there of from a list of thousands should one pick for enhanced sampling runs? How does a modeler even begin to pick starting coordinates for investigation? This remains true even in the case of simple two state systems and only increases in difficulty for multi-state systems. In this work, we solve the "initial" CV problem using a data-driven approach inspired by the filed of supervised machine learning. In particular, we show how the decision functions in supervised machine learning (SML) algorithms can be used as initial CVs ($SML_{cv}$) for accelerated sampling. Using solvated alanine dipeptide and Chignolin mini-protein as our test cases, we illustrate how the distance to the Support Vector Machines' decision hyperplane, the output probability estimates from Logistic Regression, the outputs from deep neural network classifiers, and other classifiers may be used to reversibly sample slow structural transitions. We discuss the utility of other SML algorithms that might be useful for identifying CVs for accelerating molecular simulations.

**Introduction:**
Efficient configuration space sampling of proteins or other complex physical systems remains an open challenge in computational biophysics. Despite advances in MD code bases[1], hardware, and algorithms[2], routine access to micro to millisecond timescale events is impossible for all but a few[3–7].

As an alternative to unbiased molecular simulations, enhanced sampling methods such as Metadynamics[8–11] or Umbrella sampling[12,13] offer promise. However, these require identification of a set of slow collective variables (CV) to sample along. Recently we[14–17] proposed using methods from the Markov modeling literature, namely the time-structure based independent component analysis method (tICA)[14,15,18–20] or the variational auto-encoder method from machine learning (ML) literature[16,17], for identifying optimal linear and non-linear low-dimensional CVs for accelerated sampling. However, these proposed methods are intended for a wild-type simulation that has been sufficiently sampled. While we showed[14] that full phase space convergence was not necessary for approximating such CVs, the models, and thus the CVs, became more robust as more data was input.

Alternatively, other groups[21,22] have proposed using spectral gap or re-weighting based approaches to identify appropriate CVs in an iterative fashion. However, the latter methods require effective good initial CVs since poor CVs might inadvertently introduce orthogonal modes thereby slowing convergence. In this work, we wish to answer the question: How do we select initial collective variables? There is obviously no *correct* a priori answer to this problem. Arguably, an effective starting CV should be continuously defined and differentiable, low dimensional, capable of separating out target states of interest while minimizing the number of orthogonal degrees of freedom[23]. Previous approaches for protein systems have included expertly chosen distances or dihedrals[24], path based CVs[25,26], and more generic CVs such as alpha or beta sheet character etc. Similarly, the use of ML based methods for finding CVs is not new. For example, several groups have proposed methods to use linear or non-linear dimensionality reduction based methods[8,10,27] to find low dimensional CVs. In contrast to unsupervised learning, supervised machine learning can use the state labels, enabling efficient metric learning for distinguishing between the start and end states.

Therefore, in this paper, we propose approaching the CV selection problem using an automated data driven scheme. For molecular systems *where the end states are known*, we re-cast the problem into a supervised machine learning ($SML_{cv}$) problem. SML are machine learning algorithms capable of learning a mapping between a high dimensional input to output. These algorithms are routinely used in ML to say classify pictures of cats from dogs. Similar to cats and dogs, we first show how various classification algorithms[28] from the ML literature can learn a decision boundary (Figure 1) separating *protein* states and how this enables us to define CVs for enhanced sampling. We first demonstrate this using a support vector machine ($SVM$), and how to use a configuration's distance to the decision boundary ($SVM_{cv}$) as a CV. Second, we demonstrate the same principle using a logistic regression ($LR$) classifier, and how its predicted state probabilities ($LR_{cv}$) can be used as a CV. Thirdly, we show how deep neural networks (DNNs) based classifiers produce un-normalized state probabilities ($DNN_{cv}$) that can be used as CVs. Fourthly, we show how to include data from multiple states by combining multiclass classification with multidimensional enhanced sampling. We end the paper by sampling the folding landscape of the Chignolin mini protein using $SVM_{cv}$.

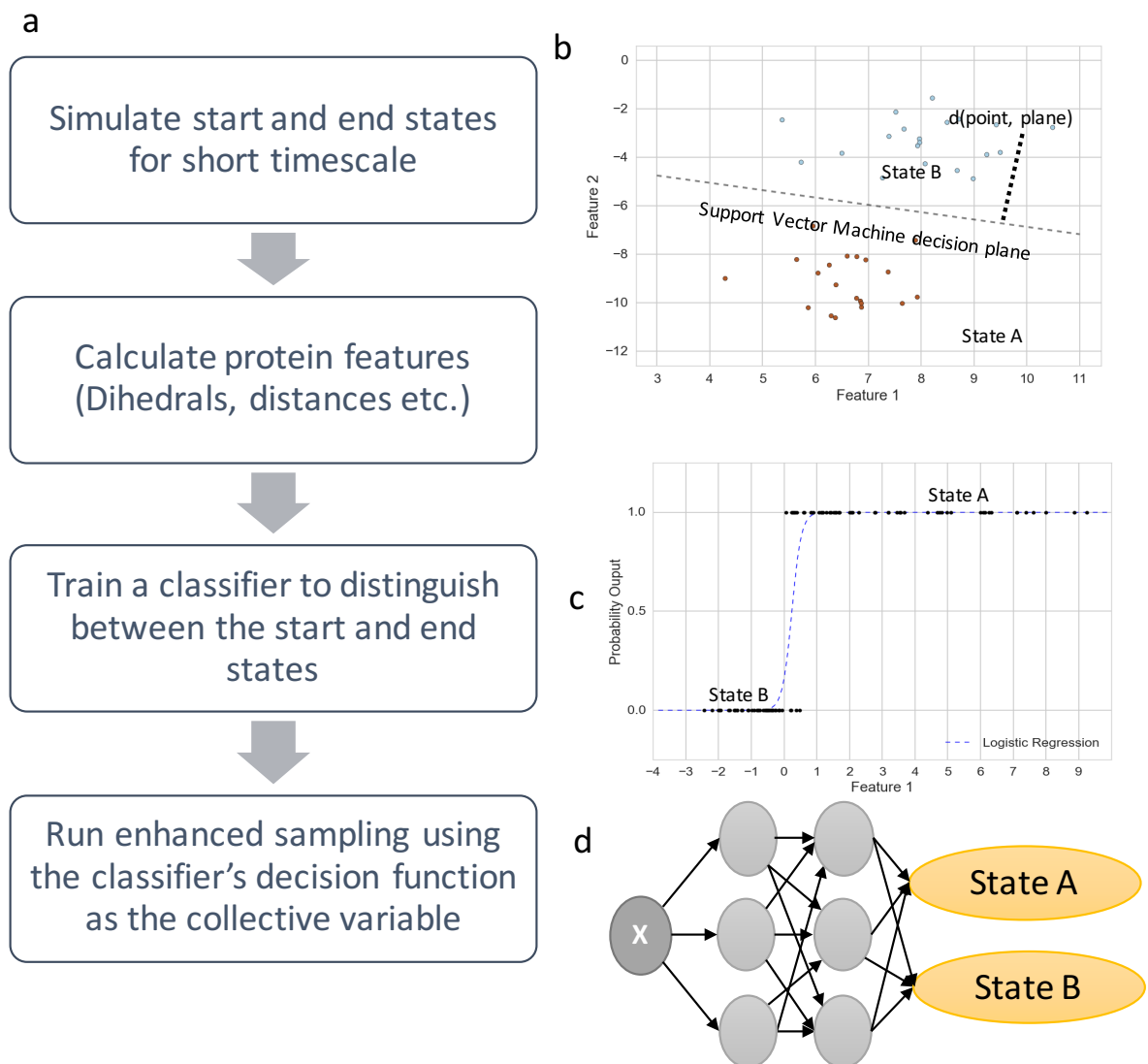

*Figure 1: Pictorial representation of the proposed method and two common classification algorithms. a). Step-wise process for generating a collective variable (CV) using supervised machine learning (SML) on known end states. b). In support vector machines (SVM), the algorithm finds a dividing hyper plane boundary between the two states of interest. The signed distance of any point to this plane becomes the collective variable. c). By contrast logistic regression (LR) classifiers models the output probability as a sigmoid transform of the weighted input features. In this instance, the predicted probability becomes the collective variable. d). In the case of the Neural network, the output of a fully connected deep network is the un-normalized probability of being in either state. In this instance, the decision function is highly non-linear function that maps the input frame X to the un-normalized output probability. Light grey circles indicate hidden layers/nodes.*

**Methods:**

All of our methods assume that the modeler has access to short trajectories from the known start and end states. After running short dynamics (on the order of ns to microseconds for protein systems) in the start (State A) and end (State B) states (Figure 1 b-c), we train a classifier to distinguish between the start and end states. To do this, we first project the trajectories onto complete list of possible order parameters/ features. For proteins, this list might include *all* backbone and side chain dihedrals, alpha carbon pseudo dihedrals, contact distances etc. An attractive advantage of using SML algorithms to build these CVs is that most of these algorithms can be explicitly regularized (via L-1 regularization for instance) to reduce model complexity. For example, if a certain dihedral or distance is incapable of distinguishing between the start and end states, then the L-1 regularizer's penalty function will allow the SML algorithm to discard that feature. More concretely, this would set the model's coefficient (see below) for that feature to be 0. In the examples below, we used an L-1 regularization for the alanine dipeptide examples and an L-2 regularization scheme for the Chignolin example. In contrast to L-1, L-2 regularization pushes all of the model coefficients towards an average value. A second benefit comes from realizing that the depositing Metadynamics bias will simultaneously accelerate *all* included degrees of freedom (Figure 2b, 4b, and 5b) which can accelerate convergence and allow for more aggressive biasing schemes. Again, more concretely, if getting from protein state A to state B requires both a dihedral and an interatomic distance to change, then using the $SML_{cv}$ would allow us to add external biasing forces to both of these features using a single collective variable. In traditional methods, the modeler would either run multi-dimensional enhanced sampling or judiciously pick and hope for the best. Multi-dimensional sampling cannot scale to more than 3 dimensions while picking and hoping scales even worse. Although, there are a wide variety of possible SML algorithms, we focus on three of most common classification methods namely SVMs, LR, and DNN classifiers.

*Support Vector Machines.* SVMs are linear classifiers that find the separating hyperplane that maximizes the distance between the closest points of the two classes (Figure 1b). Mathematically, they predict the output label as follows:

$$y = \mathbb{1}[(w^T X + b) > 0]$$

where y is the output label, X is the high dimensional input vector, $\mathbb{1}$ is the indicator function, w is the learnt vector of coefficients, and b is the scalar bias. SVM's optimization is outside the scope of this paper but it is worth noting that these models can be easily fitted via API calls to modern ML software such as scikit-learn[29]. The accompanying online IPython Notebooks[30] provide examples on how these models can be constructed.

The direct indicator function based output from SVMs is not differentiable and thus cannot be used as a collective variable for enhanced sampling. This is because a non-differentiable function of the input coordinates would have discontinuous derivative at the corresponding decision boundary. This derivative, which can be directly related to the extra forces applied to the atomic positions, would be zero everywhere inside the boundary since moving in any direction wouldn't cause a numerical change in predicted output state. However, when the particle is at the decision boundary, the derivative would rise rapidly due to a sudden change in the state. This is likely to cause the simulation to crash. However, we can use the signed closest *distance* to the SVM's hyperplane as our collective variable instead. Intuitively speaking (Figure 1b), a larger distance in either direction would mean that our current simulation frame is further in one basin or another. Thus our collective variable becomes:

$$\text{SVM}_{cv} = \text{distance}(\text{point}, \text{plane}) = \frac{(w^TX + b)}{||w||_2}$$

where we divide by $||w||_2$ because the weight vector is not normal in most circumstances. It is also possible to simply use the numerator, also called the decision value or the softness field in glass dynamics[31], as the collective variable.

*Logistic Regression.* The LR classifier (Figure 1c) models the probability of the data by maximizing the distance of each data point to the separating hyperplane.

$$P(y = 1|x) = \sigma(w^TX + b)$$

where σ is a sigmoid function:

$$\sigma(x) = \frac{1}{1 + e^{-x}}$$

While SVMs try to maximize the margin between the two classes, LR algorithms maximize the posterior probability of the classes. However, unlike SVMs, now we can directly use the differentiable probability output from the model as our collective variable for Metadynamics or other enhanced sampling algorithms.

$$\text{LR}_{CV} = P(y = 1|x) = \sigma(w^TX + b)$$

While we are using the probability of being in state 1 as our collective variable, we can also use the conditional likelihood (odds ratio) of being in either state as a collective variable as well. In that instance, the CV becomes:

$$\text{LR}_{CV} = \frac{P(y = 1|x)}{1 - P(y = 1|x)}$$

**Incorporating non-linearity via kernels or deep neural networks.**
It is entirely possible that the linear models presented above are inadequate for separating the training simulation given the feature space. This can be diagnosed via hysteresis during the enhanced sampling simulations. In that case, we recommend two possible extensions. 1) Using a kernel function[32,33] to implicitly encode the input feature vector unto a high dimensional space. However, it is worth noting that kernel functions often require picking landmark[19,32,34] protein configurations for efficient prediction and preventing over-fitting. 2) Alternatively, non-linearity can be achieved via deep neural networks[15] (DNNs). DNNs are universal function approximators. Typical DNNs consist of a series of fully connected affine transformation layers (Figure 1d) interspersed with non-linear activation function, such as the Sigmoid, ReLU, Swish[35]. Previous works have already highlighted the expressive power of neural networks for dimensionality reduction[16,36] and sampling[17,37]. Here we argue that given some labeled state/trajectory data, the un-normalized output from these networks could now be used as a set of differentiable collective variables for accelerating molecular simulations.

$$\text{DNN}_{\text{CV}} = \text{Unnormalized Deep Neural Network Output} = G(X)$$

Where $G(X)$ is the non-linear transformation learnt by the DNN to accurately separate the training examples. However, we caution users against constructing arbitrarily deep networks since they might be difficult to train or make the force calculations computationally expensive, thereby negating the speed ups from accelerated sampling.

**Extension to multiple states via bias-exchange:**
Up to now we have only considered the problem of generating a collective variable for a two state system. However, most large biophysical[3,4] and non-biological systems have multiple metastable states. In several instances, theses states are known before hand from crystallography or NMR data but the relative free energies or kinetics of inter-conversion between those states are not known. To incorporate these states into our model, we recommend generating a multiclass classification[28] model. Given N states, the output from our SML algorithm would now be a N-dimensional probability vector, and we can use the multiple outputs as N individual coordinates in a bias-exchange simulation[38] or accelerate them simultaneously using multi-dimensional Gaussians (if the number of states is <3). While certain algorithms, such as DNNs, naturally allow for prediction of multiple outputs, other SML algorithms adopt a "one-vs-rest" (OVR) strategy. In OVR, we build a series of models for each output state such that the model learns a hyper boundaries or decision functions that separate the current target state from all the other states. Thus we get N sub-models for the N states, and given a new data point, we compute its distance to each of those hyper planes and assign it to its closest state. However, in-lieu of prediction, we can now use the same *set* of *decision functions as individual collective variables* in an enhanced sampling simulation. For example, in the case of a 3-state SVM, the 3 collective variables would be :

$$\text{CV}_1 = \text{State 1 vs rest} = \text{distance}(\text{point}, \text{State 1 hyper plane}) = \frac{(w_1^T X + b_1)}{||w_1||_2}$$

$$\text{CV}_2 = \text{State 2 vs rest} = \text{distance}(\text{point}, \text{State 2 hyper plane}) = \frac{(w_2^T X + b_2)}{||w_2||_2}$$

$$\text{CV}_3 = \text{State 3 vs rest} = \text{distance}(\text{point}, \text{State 3 hyper plane}) = \frac{(w_3^T X + b_3)}{||w_3||_2}$$

Where $w_1, w_2,$ and $w_3$ and $b_1, b_2,$ and $b_3$ are the respective weight vectors and biasing scalars for the hyper planes that separate their respective states from the rest of the ensemble. Similar expressions can be derived for other SML methods such as Logistic regression classifiers. On the other hand, since DNNs can directly output an un-normalized probability vector of length equal to the number of states, there is no need to use OVR based strategies.

**Notes on implementation and model cross-validation**

It is worth recognizing that using this SML based CVs requires existing enhanced sampling software to have native machine learning libraries built in so that the collective variable definitions proposed above can be utilized. However, at their core, these machine learning models are nothing more than a series of vector operations whose compute graph can be easily broken down and implemented in a step wise fashion in most modern enhanced sampling engines. To highlight this, our open source implementation at the end of the manuscript provides examples on how the SVM, multiclass-SVM, LR, or even DNNs can be written as a series of custom scripts such that enhanced sampling packages such as Plumed and OpenMM[39,1] can interpret and utilize the mathematical transforms that connect the current simulation frame's atomic coordinates to a set of scalars—aka the collective variables.

To prevent over fitting, we also recommend k-fold (k=3-10) cross-validation to identify optimal model hyper parameters. In cross-validation, the model is trained on a subset of the data (the training data), and its accuracy is scored on a held out set (validation data). A range of models with varying hyper-parameters and model complexity are built, and parameters for the best scoring model are saved. The best model can then be retrained on the full training and validation sets before reporting its performance on a held out test set. Since the "test" for our model is its ability to accelerate sampling, we chose to simply do 3-fold cross-validation without a held out test set. The Supporting Information contains validation set accuracies for a range of different models and systems. In general, across several reasonable parameter values (Supporting Figure 3-5), our models' accuracies never drop below .92 (92%) for either example. For alanine (Supporting Figure 3-4), all models reported accuracies of 100% because the alpha and beta regions are easily linearly and non-linearly separable. As specified above, after cross-validation, we re-trained the models on the full train and validation dataset.

**Results:**

**Application to alanine dipeptide using SVMs**

We showcase our methods for three different linear classifiers on solvated alanine dipeptide. The training simulations were previously generated[14], and consisted of two 2ns trajectories starting from the $\beta$ and $\alpha_L$ basins on the Ramachandran plot (Figure 2). Similar to previous work, we used the sin-cosine transform of the backbone $\phi$ and $\psi$ dihedrals[19,40] as input to our models. Thus each training simulation frame was represented as 4 numbers. The control simulations (Supporting Figure 1-2) were accelerated only along the backbone $\phi$ and $\psi$ dihedral. All models were trained using scikit-learn[29] or PyTorch[41]. For the SVM and LR models, we performed 3-fold cross-validation to determine the best hyper-parameters and regularization strengths (Supporting Figures 3-4) but found that our dataset was simple enough that a large range of hyper parameter settings gave very similar results. Ultimately, we picked reasonable model parameters (Supporting Figure 3-4) and then re-trained the model on the entire available 4ns. For the SVM and the LR models, we used a L-1 regularization with a the regularization strength (C-parameter) of 1.0. The Supporting Information contains the full model parameters and the online Github repository contains the fitted models. The model was trained once and kept fixed throughout the simulation. The pre-fitted models were written as custom scripts to perform Metadynamics using Plumed[39] and OpenMM[1]. The Metadynamics simulations were well-tempered[42] with the parameters shown in Supporting Table 1. The simulations were started from the $\beta$ basin, run for a total length of 45 ns apiece, and saved every 10ps. All other simulation parameters were kept the same as before[14]. The trajectories were analyzed and plotted using standard python libararies[19,29,43,30] while reweighting was done using the time-independent estimator of Tiwary[11].

The results from our $SVM_{cv}$ simulations are shown in Figure 2. Based upon the 4ns of training data (Figure 2a), the SVM model is able to find a hyperplane separating the $\beta$ and $\alpha_L$ regions (Figure 2a). The distance to this hyper plane (Figure 2a color bar) now becomes our CV for accelerating sampling via Metadynamics (Figure 2c). Over a 45ns run, we sample the slowest coordinate $\beta$ to $\alpha_L$ transition tens of times, allowing for robust estimation of the free energy surface via reweighting (Figure 2d). Additionally, since the simulations were converged, we are able to efficiently sample the faster orthogonal $\psi$ degree of freedom as well. Similar to the SVM, we also used the Logistic Regression Classifiers (LRs) to define CVs ($LR_{cv}$) for accelerated simulations. The Supporting Note 1 and Supporting Figure 6 contain details on training and using these models for defining CVs. We recommend modelers start with simpler linear models before moving to the more complex DNN based approaches shown next.

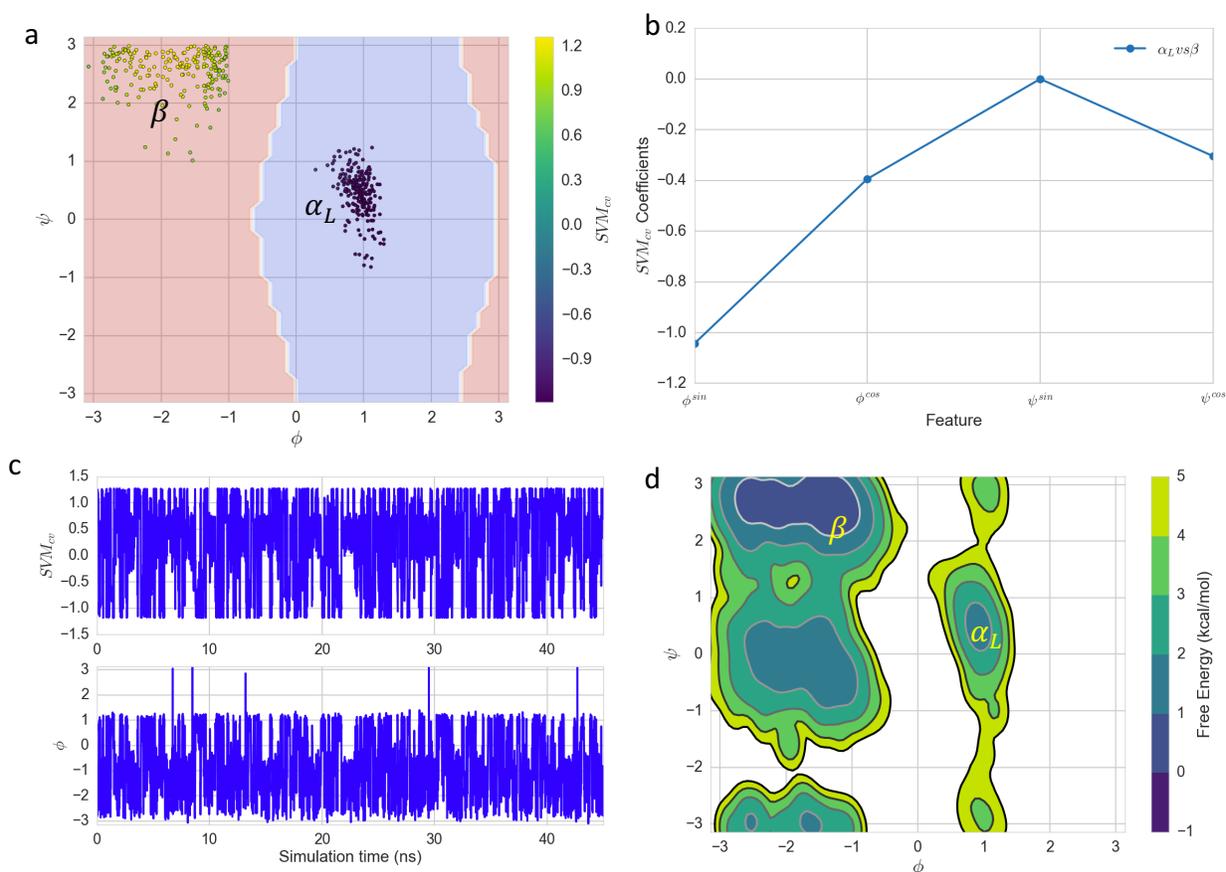

*Figure 2: Results from $SVM_{cv}$ based sampling. a) The two 2ns training trajectories projected unto the Ramachandran plot. Note that no transition was observed from the $\beta$ to the $\alpha_L$ basins. The training frames are colored according to the frame's distance to the SVM's decision boundary in 4 dimensional feature space. The contours show the decision boundary for each region. b) Decomposing the SVM's coefficients vector into the original training features shows the model assigns higher weights to the $\phi$ features than the $\psi$ features. The L-1 regularization drops the cosine transform of the $\psi$ backbone dihedral to induce sparsity. c)*

*Running well-tempered Metadynamics simulations along the $SVM_{cv}$ efficiently samples the β to $α_L$ transition repeatedly. d) Reweighting allows us to recover the full free energy surface along the Ramachandran plot.*

**Application to alanine dipeptide using DNNs**

We next looked into using deep neural networks (DNNs) to find non-linear boundaries between the start and end states. To show this, we trained a simple 5-layered DNN and used the resulting model as a collective variable for enhanced sampling via Plumed and OpenMM[1,39]. The DNN's layers alternated (Figure 3b) between fully connected affine transformations and non-linear activation functions. For the non-linear activation layer, we used the recently developed Swish[35] non-linear function.

We trained the model (Figure 3b) on half of the 4ns alanine dipeptide dataset using PyTorch. We note that the output layer of this model gives two values which are the un-normalized probability of the input frame being in either state. Either (or both) of these output nodes can be used as a collective variable since, once normalized, they sum up to 1. To train the model, we minimized the cross-entropy between the models' output and the training data. This was done using the Adam optimizer[44] with an initial learning rate of 0.1. We trained the model for 1 epoch using batch size of 32. The optimization was performed using Pytorch on the CPU platform and training took less than a minute. After 1 epoch, our model's training error was below 0.1 and model reported 100% accuracy on a held out 2ns test set. Therefore, we stopped training at this instance, and used this model going forward. Once the model had been trained (Figure 3a), we wrote custom string expressions to convert the Pytorch DNN model into a format that Plumed can process. Again, the Metadynamics simulations were well-tempered[42] with the parameters shown in Supporting Table 1. The online Github repo provides examples on how such models can be trained and transferred to Plumed.

The results are in Figure 3c-d, running Metadynamics simulations along the output probability, $DNN_{cv}$, allows us to observe >15 transitions (Figure 3c) along alanine's slower $\phi$ coordinate in just 45ns of sampling. Similar to the SVM model, the robust transition statistics combined with the reweighting method of Tiwary[11] now allow us to recover the free-energy landscape (Figure 3d) across other coordinates of interest.

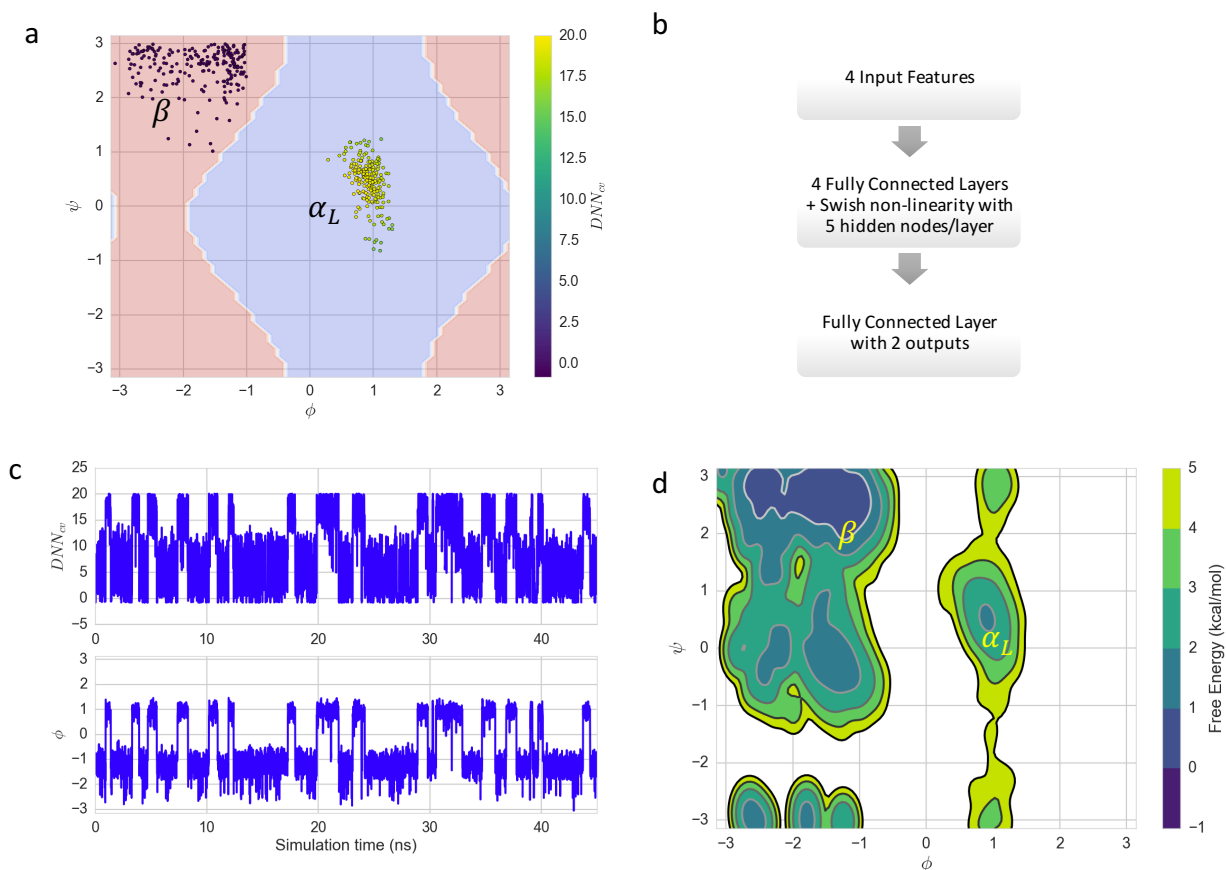

*Figure 3: Results from $DNN_{CV}$ based sampling. a) The training and test trajectories projected unto the Ramachandran plot. Note that no transition was observed from the β to the $α_L$ basins. The frames are colored according to the unnormalized output of the DNN for frame being in the $α_L$ state. The contours show the decision boundary for each region. b) Architecture of the neural network used in this study. We didn't optimize the architecture but did perform a 50-50 training and test split. The two outputs correspond to the un-normalized probability of being in β or $α_L$ basin. Either of the output (or both of them) can be used as a collective variable. In this instance, we used the output corresponding to the un-normalized output of the DNN for the frame being in the $α_L$ state. c) Running well-tempered Metadynamics simulations along the $DNN_{cv}$ efficiently samples the β to $α_L$ transition repeatedly. d) Reweighting allows us to recover the full free energy surface along the Ramachandran plot.*

**Application to alanine dipeptide using multiclass supervised machine learning**

We next extended our method to the multiple state scenario. This is necessary because most large systems such as kinases, GPCRs, or ion channels exhibit multi-state characteristics that are often available as crystallographic starting points[4]. Simulations around these local points can be used to define a set of decision functions that separate each state from the rest. This set of decision functions can then be used in a multi-dimensional enhanced sampling run.

To test the method, we generated three 1 ns trajectories in the β, $α_L$ and $α_R$ region of alanine dipeptide (Figure 3a). Based off the cross-validation results in the previous 2-state SVM model, we built an L-1 regularized 3-state SVM with the regularization strength set to 1.0 and trained it using a squared-hinge loss (Supporting Information). The model parameterizes 3 hyper-planes, and the distance to each of those hyper planes was used in a 3-dimensional Metadynamics simulation. It is trivial to separate these into three one dimensional Metadynamics simulations, connected via Hamiltonian replica exchange[38]. All Metadynamics parameters were the same except that we used last-bias reweighting, in place to Tiwary's time-dependent estimator, due to numerical stability issues. We limited our simulations to 12ns because the simulation showed excellent convergence (Figure 4c below) well before that.

The resulting multiclass SVM model (Figure 4a) learnt a set of dividing boundaries or functions that separate each of $\beta$, $\alpha_L$ and $\alpha_R$ region from the other two. The coefficients for the respective hyper planes are shown in Figure 4b. For example, as intuitively expected, $\alpha_L$ plane (Figure 4b orange) is separated from the rest of the ensemble only via the $\phi$ dihedral features. In this instance, regularization forces the other two coefficients to 0. By contrast, $\alpha_R$ plane has non-zero weights on both $\phi$ and $\psi$ features. The simulations quickly reach the diffusive regime (Figure 3c), showing tens of transitions in just 10ns and highlighting the effectiveness of the multiclass SML algorithm as a set of collective variables. Last bias reweighting gives us a similar free-energy surface as before (Figure 3d).

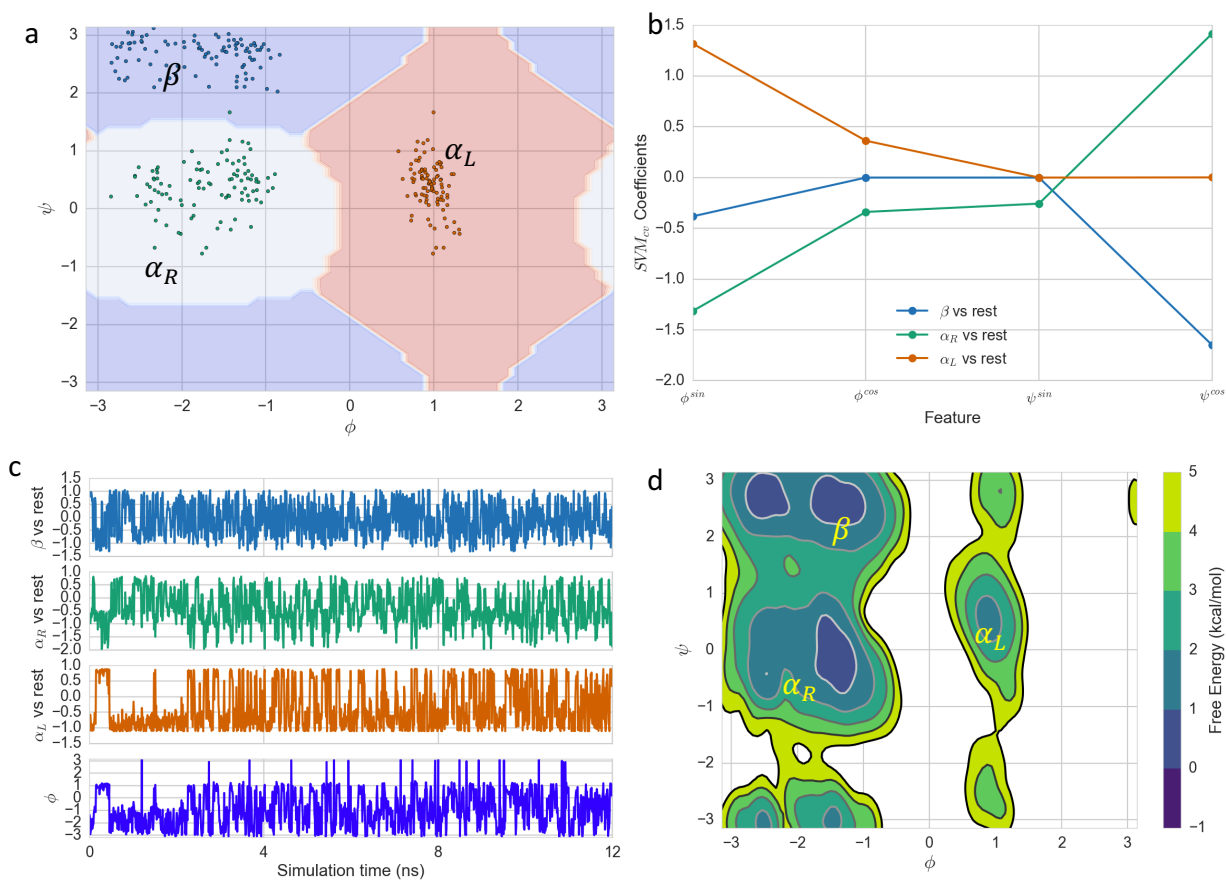

Figure 4: Results from multiclass $SVM_{cv}$ based sampling. a) The three 1ns training trajectories projected unto the Ramachandran plot. Note that no transition was observed in either dimension. The contours show the decision boundary for each region. b) Decomposing the SVM's coefficients vector into the original training features shows the model assigns higher weights to the $\phi$ features than the $\phi$ features for the $\alpha_L$-vs-rest hyper plane but assigns higher weights to the $\psi$ dihedrals for the other two. Several

*coefficients are dropped due to regularization. c) Running well-tempered Metadynamics simulations along the $SVM_{cv}$ efficiently samples the $\beta$ to $\alpha_L$ transition repeatedly. d) Reweighting allows us to recover the full free energy surface along the Ramachandran plot.*

To compare our simulations to a baseline, we ran two control 12+ns simulation (Supporting Figure 1) where we accelerated the dynamics of the $\psi$ or $\phi$ dihedral alone. $\psi$ is a bad starting CV since the slowest transitions in alanine are dominated by the $\phi$ dihedral. Thus, in our first control simulation, we only observed three transitions along the $\phi$ plane. By contrast, both variants of $SML_{cv}$ were able to observe at least 5 transitions in the same amount of simulation time. Additionally, all simulations, including the control, were also able to drive the slow uphill $\beta$ to $\alpha_L$ transition much more quickly (<5ns) than the unbiased 43ns mean first passage time estimate previously reported[45]. However, it is worth noting that some of our SML based CV are not as efficient as simply directly accelerating the $\phi$ dihedral (Supporting Figure 2). This likely indicates that we do not have the most efficient possible CV, but it is entirely possible to now turn our initial SML based CV into a better CV by combining it with the SGOOP formalism[21]. Intriguingly, knowledge about the $\phi$ could also potentially be incorporated as Bayesian prior into the $SML_{CV}$ formalism.

**Application to protein folding using SVMs**

We next sought to fold the 10 residue Chignolin mini-protein[46] using the Amber forcefield[47], TIP3P[48] water model, and vanilla NVT simulations. To that end, we obtained the all atom Chignolin trajectories from D.E. Shaw research[46,49]. We then sliced 1000 frames (2 $\mu s$ trajectory) from the folded and unfolded (Figure 5a) states (4 $\mu s$ total) respectively, ensuring that no actual folding event (Figure 5a) was seen in those trajectories. To featurize these 2000 frames, we used a combination of the sin-cosine transform of the backbone $\phi, \psi$ dihedrals, $\alpha$ carbon contact distances for residues at least 3 apart in sequence, and the cosine transform of four consecutive $\alpha$ carbon atoms. Thus each frame in our simulation was represented using 78 features. We normalized these features to have close to zero mean and unit variance using a Standard scalar trained on the Shaw simulations[7,46]. This scaling is necessary because most SML algorithms have issues with varied feature scaling—for example dihedrals going from negative pi to pi while distances going up to tens of angstroms. Decision trees and Random Forests do not require this feature scaling because unlike SVMs, LR, or DNNs they do not learn a "distance" metric and are invariant to monotonic feature scaling.

We trained a SVM (Figure 5b) on these 2000 frames (1000 folded and 1000 unfolded), using cross-validated parameters (Supporting Figure 5), via Scikit-learn[29]. For this model, we used L-2 regularization with the regularization strength set to 1.0. However, changing the regularization by a factor of 10 in either direction had minimal effect on the final decision boundary (Supporting Figure 5). Then similar to the alanine example, we used the distance to the SVM's hyper plane as our collective variable. In this instance, we started 25 Metadynamics using well-tempered metadynamics[9,42]. Each of our walkers was run for 100ns (~18hrs of parallelized sampling on commodity K40 GPUs). Thus the total amount of sampling was 2.5 µs. We ran vanilla NVT Metadynamics simulations using a 2fs Langevin integrator with a friction coefficient of 1/ps , and the estimated melting temperature of 340K[46]. The Metadynamics simulations were well-tempered[42] with an initial height of 1kJ/mol, a bias factor of 8, and deposition rate of 2ps. The sigma parameter of 0.2 was chosen based off the training simulation data in the folded state. The simulations frames were saved every 50ps. See Supporting Table 2 for the rest of the Metadynamics parameters.

Over the 100ns simulations, at least 5 out of 25 walkers naturally folded to the correct Chignolin topology with Figure 5c and Figure 5d insets showing representative traces and structures. The supporting movie shows an example of such a folding event. Clearly, our $SVM_{cv}$ is able to efficiently drive the simulations between the folded and unfolded ensemble since we are able to obtain multiple folding events in total simulation time comparable to the starting training simulation in a single basin. We also used the time-independent free energy estimator[11] to re-weight the simulations (Figure 5d), showing, similar to previous works[46,50], that the folded state was the dominant free energy state. Unfortunately, the flattened barrier (Figure 5d) likely indicates that the transition states were flattened during the course of our simulations, making kinetic estimates difficult with these simulations.

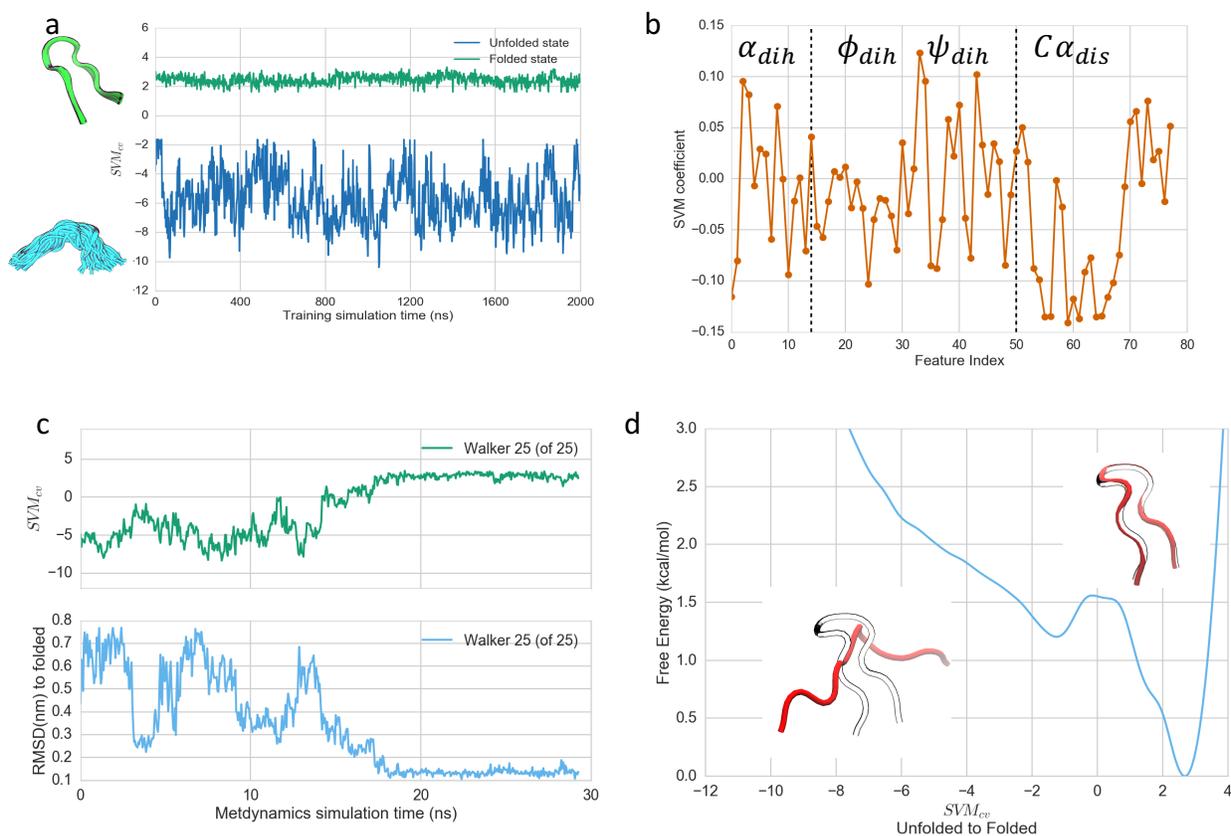

*Figure 5: Results from $SVM_{cv}$ based sampling on the Chignolin ten-residue mini-protein. a) The two 2µs training trajectories projected unto the learnt decision function. Note that no transition was observed from the folded to the unfolded basin. The protein images on the left show several randomly sampled frames. b) The $SVM_{cv}$ decision function is a complex linear combination of dihedrals and contacts. Note that we used normalized contacts and dihedrals as inputs to the ML model. c) One of several folding trajectories obtained from multiple walker well-tempered Metadynamics using the $SVM_{cv}$. The top panel traces the collective variable while the bottom traces the root mean squared deviation of the alpha carbons to the NMR structure 2RVD[51]. d) Integrating the hills, and reweighting[11] allows us to construct the free energy profile. The insets show randomly selected*

*unfolded and folded structures in red with the NMR structure in white. This image was generated using IPython notebook[30], MSMBuilder[19], MSMEplorer[43], and VMD[52].*

**Conclusion:**
We believe the power of our method $SML_{cv}$ comes in automating the CV selection problem, providing a systematic way to include non-linearity and regularization, and minimizing the work that needs to be done to find starting points for enhanced sampling runs. Our results further open up significant opportunities for both benchmarking existing SML algorithms, and developing novel algorithms or loss functions inspired from statistical mechanics. For example, other SML techniques, such as the naïve Bayes classifier, might present interesting starting points for building classification schemes better suited for accelerated sampling. However, not all SML algorithms are amenable to being used as CVs. For example, while decision trees and random forest classifiers[53,54] are useful for understanding molecular systems[54], their indicator function based optimization procedure is not differentiable, making their decision function unlikely to be used as a CV.

Our proposed method ($SML_{cv}$) is intended to produce a first estimate of CVs from very limited data collected from locally sampling each basin, and as such will likely not be able to generate an ideal CV. In all likelihood, the model might accidently induce orthogonal degrees of freedom in either state thereby slowing convergence. However, this problem would exist for any of the existing methods in literature.

While we have focused on pre-training and generation of CVs, we also believe that supervised machine learning based collective variables ($SML_{cv}$) might be excellent starting coordinates for further optimization via SGOOP[21] or VAC[22]. This would make the entire process into an online learning setup where each additional round of simulation improves the dividing hyper boundary. Additionally, these coordinates are likely to be transferable[15,17] across related systems which would make them useful for investigating drug binding/unbinding kinetics, mutational effects, modest force field effects etc. However, when and where transfer learning might fail is an unsolved problem. Lastly, we hypothesize that the automated order parameter identification might make it easier to converge chain-of-states based methods such as nudged elastic band or string based method[25,55–57] since the separating hyper plane can now be used to more efficiently guide the iterative minimum free energy path finding algorithms—similar to kernel-SVM approach of Pozun et al[58]. We hope that these results provide stimulating starting points for automatic CV optimization protocols, allowing modelers to focus more on the results of free-energy simulations rather than their initial design.

**Supplementary Material:**
The supplementary material contains several additional figures which are comparisons of the $SML_{cv}$ results against hand picked collective variables for alanine dipeptide, results of 3-fold cross-validation for the given models, and results from using a Logistic regression classifier's decision function as a collective variable for accelerated sampling.

**Acknowledgements:**

The authors would like to thank various members of the Pande lab for useful discussions and feedback on the manuscript. M.M.S. would like to acknowledge support from the National Science Foundation grant NSF-MCB-0954714. This work used the XStream computational resource, supported by the National Science Foundation Major Research Instrumentation program (ACI-1429830). The authors would like to thank D.E. Shaw and DESERES for graciously providing the Chignolin folding trajectories. VSP is a consultant and SAB member of Schrodinger,LLC and Globavir, sits on the Board of Directors of Apeel Inc, Freenome Inc, Omada Health, Patient Ping, Rigetti Computing, and is a General Partner at Andreessen Horowitz.

**Supporting videos:**

Due to size restrictions the supporting video have been uploaded online: https://www.youtube.com/watch?v=jgor4xbAai8 and https://youtu.be/D1cXDKrHMCA

**Code and data availability:**

All the models and code needed to reproduce the main results of this paper are available at https://github.com/msultan/SML_CV/

**2006**, *110* (8), 3533–3539.
(10) Abrams, C.; Bussi, G. Enhanced Sampling in Molecular Dynamics Using Metadynamics, Replica-Exchange, and Temperature-Acceleration. *Entropy* **2013**, *16* (1), 163–199.
(11) Tiwary, P.; Parrinello, M. A Time-Independent Free Energy Estimator for Metadynamics. *J. Phys. Chem. B* **2015**, *119* (3), 736–742.
(12) Kästner, J. Umbrella Sampling. *Wiley Interdiscip. Rev. Comput. Mol. Sci.* **2011**, *1* (6), 932–942.
(13) Jo, S.; Suh, D.; He, Z.; Chipot, C.; Roux, B. Leveraging the Information from Markov State Models to Improve the Convergence of Umbrella Sampling Simulations. *J. Phys. Chem. B* **2016**, *120* (33), 8733–8742.
(14) Sultan, M. M.; Pande, V. S. TICA-Metadynamics: Accelerating Metadynamics by Using Kinetically Selected Collective Variables. *J. Chem. Theory Comput.* **2017**, *13* (6), 2440–2447.
(15) Sultan, M. M.; Pande, V. S. Transfer Learning from Markov Models Leads to Efficient Sampling of Related Systems. *J. Phys. Chem. B* **2017**, acs.jpcb.7b06896.
(16) Hernández, C. X.; Wayment-Steele, H. K.; Sultan, M. M.; Husic, B. E.; Pande, V. S. Variational Encoding of Complex Dynamics. *arXiv Prepr.* **2017**, arXiv:1711.08576.
(17) Sultan, M. M.; Wayment-Steele, H. K.; Pande, V. S. Transferable Neural Networks for Enhanced Sampling of Protein Dynamics. *arXiv Prepr.* **2018**, arXiv:1801.00636.
(18) Pande, V. S.; Beauchamp, K.; Bowman, G. R. Everything You Wanted to Know about Markov State Models but Were Afraid to Ask. *Methods* **2010**, *52* (1), 99–105.
(19) Harrigan, M. P.; Sultan, M. M.; Hernández, C. X.; Husic, B. E.; Eastman, P.; Schwantes, C. R.; Beauchamp, K. A.; Mcgibbon, R. T.; Pande, V. S. MSMBuilder: Statistical Models for Biomolecular Dynamics. *Biophys. J.* **2016**, *112* (1), 10–15.
(20) Pérez-Hernández, G.; Paul, F.; Giorgino, T.; De Fabritiis, G.; Noé, F.; Perez-hernandez, G.; Paul, F. Identification of Slow Molecular Order Parameters for Markov Model Construction. *J. Chem. Phys.* **2013**, *139* (1), 15102.
(21) Tiwary, P.; Berne, B. J. Spectral Gap Optimization of Order Parameters for Sampling Complex Molecular Systems. *Proc. Natl. Acad. Sci. U. S. A.* **2016**, *113* (11), 2839–2844.
(22) McCarty, J.; Parrinello, M. A Variational Conformational Dynamics Approach to the Selection of Collective Variables in Metadynamics. *J. Chem. Phys.* **2017**, *147* (20), 204109.
(23) Fiorin, G.; Klein, M. L.; Hénin, J. Using Collective Variables to Drive Molecular Dynamics Simulations. *Mol. Phys.* **2013**, *111* (22–23), 3345–3362.
(24) Lovera, S.; Sutto, L.; Boubeva, R.; Scapozza, L.; Dölker, N.; Gervasio, F. L. The Different Flexibility of c-Src and c-Abl Kinases Regulates the Accessibility of a Druggable Inactive Conformation. *J. Am. Chem. Soc.* **2012**, *134* (5), 2496–2499.
(25) Pan, A. C.; Sezer, D.; Roux, B. Finding Transition Pathways Using the String Method with Swarms of Trajectories. *J. Phys. Chem. B* **2008**, *112* (11), 3432–3440.
(26) Meng, Y.; Lin, Y. L.; Roux, B. Computational Study of the "DFG-Flip" conformational Transition in c-Abl and c-Src Tyrosine Kinases. *J. Phys. Chem. B* **2015**, *119* (4), 1443–1456.
(27) Ceriotti, M.; Tribello, G. A.; Parrinello, M. Demonstrating the Transferability and the Descriptive Power of Sketch-Map. *J. Chem. Theory Comput.* **2013**, *9* (3), 1521–1532.
(28) Tan, P.-N. Classification: Basic Concepts, Decision Trees and Model Evaluation. *Introd. to Data Min.* **2006**, 145–205.
16

Supporting Information

**Model parameters:**
All of our SVM and LR models were generated using scikit-learn. The DNN model was generated using Pytorch. The following parameters were used for each of the models.

**Alanine 2-state SVM model (Figure 2):**
LinearSVC(C=1, class_weight=None, dual=False, fit_intercept=True,
   intercept_scaling=1, loss='squared_hinge', max_iter=1000,
   multi_class='ovr', penalty='l1', random_state=None, tol=0.0001,
   verbose=0)

**Alanine DNN model (Figure 3):**
   See the main manuscript for full details.

**Alanine 3-state SVM model (Figure 4):**
LinearSVC(C=1, class_weight=None, dual=False, fit_intercept=True,
   intercept_scaling=1, loss='squared_hinge', max_iter=1000,
   multi_class='ovr', penalty='l1', random_state=None, tol=0.0001,
   verbose=0)

**Alanine LR model (SI Figure 6):**
LogisticRegression(C=1.0, class_weight=None, dual=False, fit_intercept=True,
   intercept_scaling=1, max_iter=100, multi_class='ovr', n_jobs=1,
   penalty='l1', random_state=None, solver='liblinear', tol=0.0001,
   verbose=0, warm_start=False)

**Chignolin SVM model (Figure 5):**
SVC(C=1.0, cache_size=200, class_weight=None, coef0=0.0,
   decision_function_shape=None, degree=3, gamma='auto', kernel='linear',
   max_iter=-1, probability=False, random_state=None, shrinking=True,
   tol=0.001, verbose=False)

| Parameter | Value |
| --- | --- |
| **Gaussian Height** | 1 kj/mol |
| **Gaussian width** | 0.5(DNN), 0.1(SVM), 0.01(LR) |
| **Bias Factor** | 8 |
| **Gaussian Drop rate** | 2ps |
| **Feature Space** | Dihedrals + SinCos Transform |
| **Normalized Features** | False (all features have similar scale) |

Table 1: Set of parameters used for the SML-Metadynamics simulations of Alanine dipeptide across 3 different CV.

| Parameter | Value |
| --- | --- |
| **Gaussian Height** | 1 kj/mol |
| **Gaussian width** | 0.2 (SVM based CV) |
| **Walkers** | 25 |
| **Walkers read rate** | 50ps |
| **Bias Factor** | 8 |
| **Gaussian Drop rate** | 2ps |
| **Simulation Temp.** | 340K |
| **Feature Space** | Dihedrals(sin-cosine transform) + alpha contacts + alpha carbon dihedrals(sin-cosine transform) |
| **Normalized Features** | Yes |

Table 2: Set of parameters used for the SML-Metadynamics simulations of Chignolin

The fitted models and code are available as pickle files on the following github page: https://github.com/msultan/SML_CV/

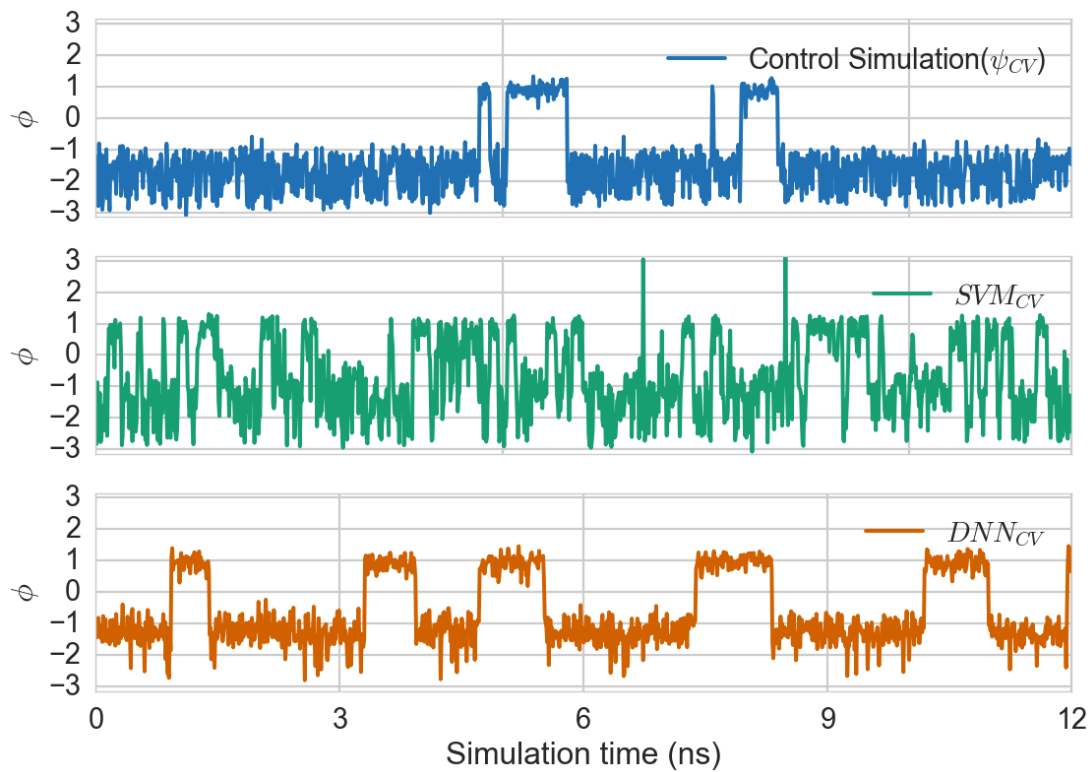

Figure 1: Comparison of $SML_{cv}$ against a badly chosen collective variable $\psi$. All three graphs trace the evolution of the slower $\phi$ dihedral over the course of the simulation. The top graph is the control simulation in which $\psi$ was accelerated while the bottom two graphs are reproduced from the main text. In contrast to a badly chosen CV, our automatic CV method is able to more efficiently drive the simulation from one state to another.

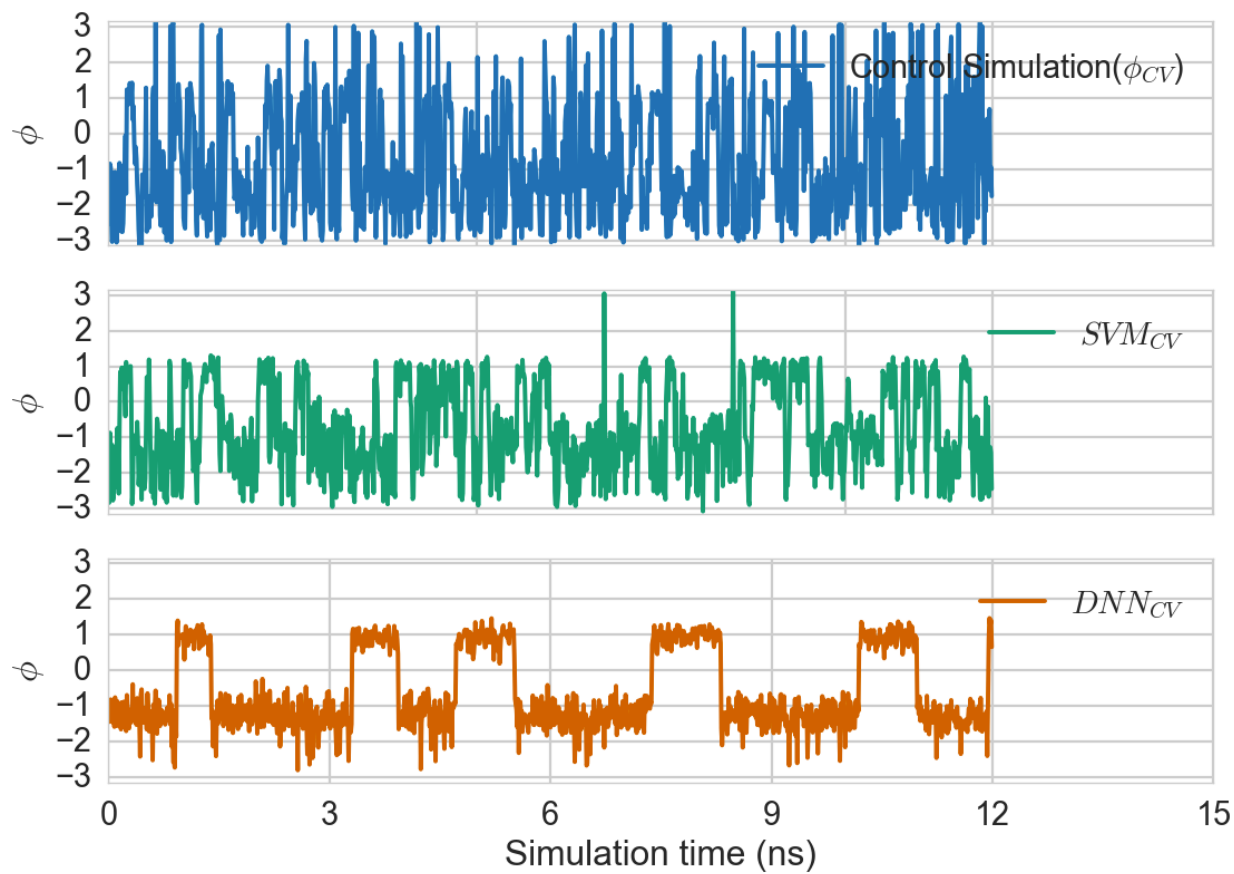

Figure 2: Comparison of $SML_{cv}$ against a good collective variable $\phi$. All three graphs trace the evolution of the $\phi$ dihedral over the course of the simulation. The top graph is the control simulation in which $\phi$ was accelerated while the bottom two graphs are reproduced from the main text. In contrast to an expertly chosen CV, our automatic CV method is less efficient though it is worth noting that the $SML_{cv}$ requires no prior knowledge about the system.

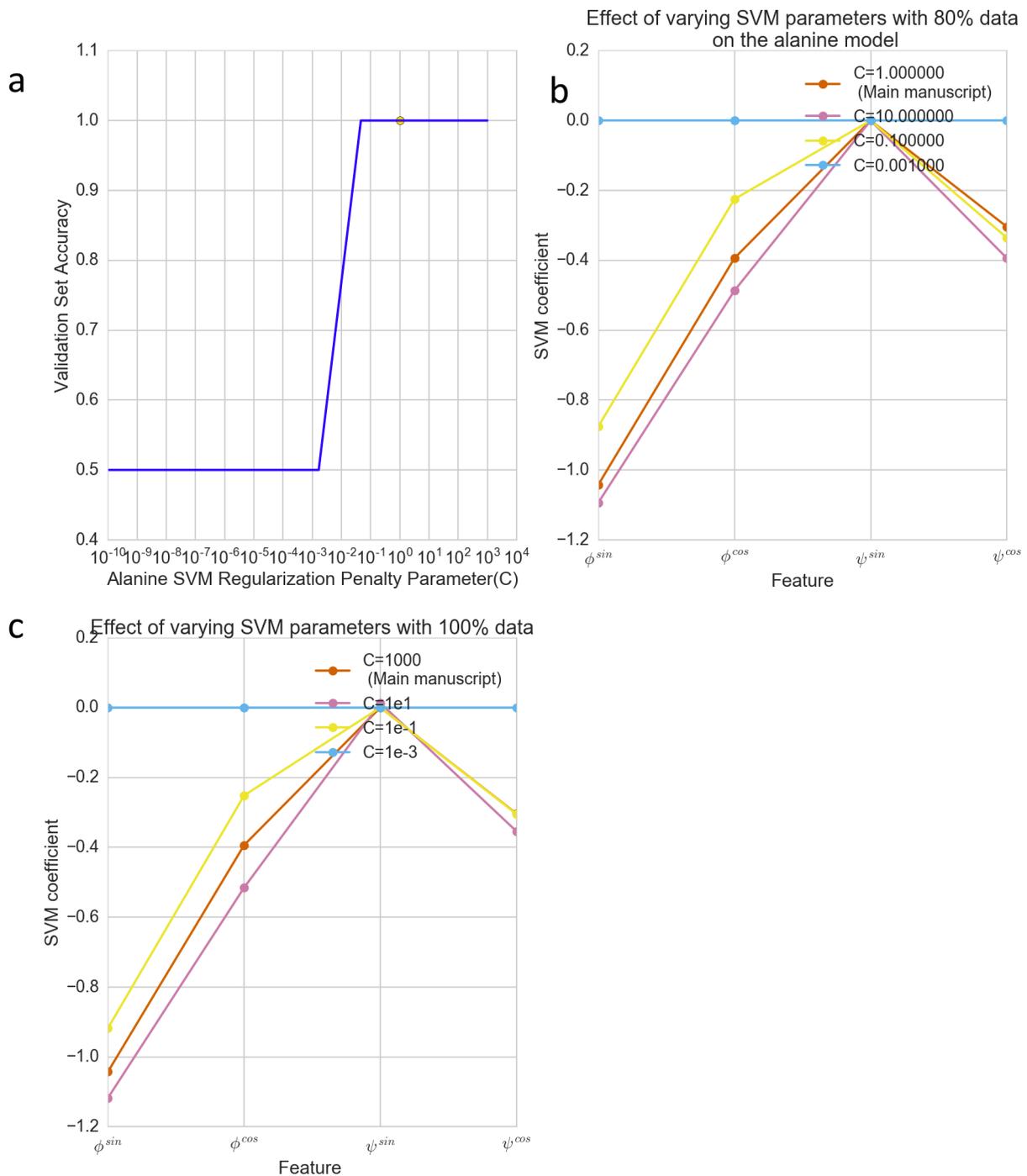

Figure 3: Sensitivity analysis for the L-1 regularized alanine SVM model. a). Results from 3-fold cross-validation shows that the model has a very high validation accuracy (~1) across several magnitude changes in the C-parameter. The error bars are negligible in the validation dataset (n=3). The gold circle is the model presented in the paper after it has been fit to the complete dataset. b). Changing the SVM's L1 regularization/slack parameter by orders of magnitude has minimal effect on the SVM coefficient. In this case, we randomly removed 20% of the data to show that the model was still robust to perturbations. c). This result remains consistent with 100% of the data across several magnitude changes in the C-parameter.

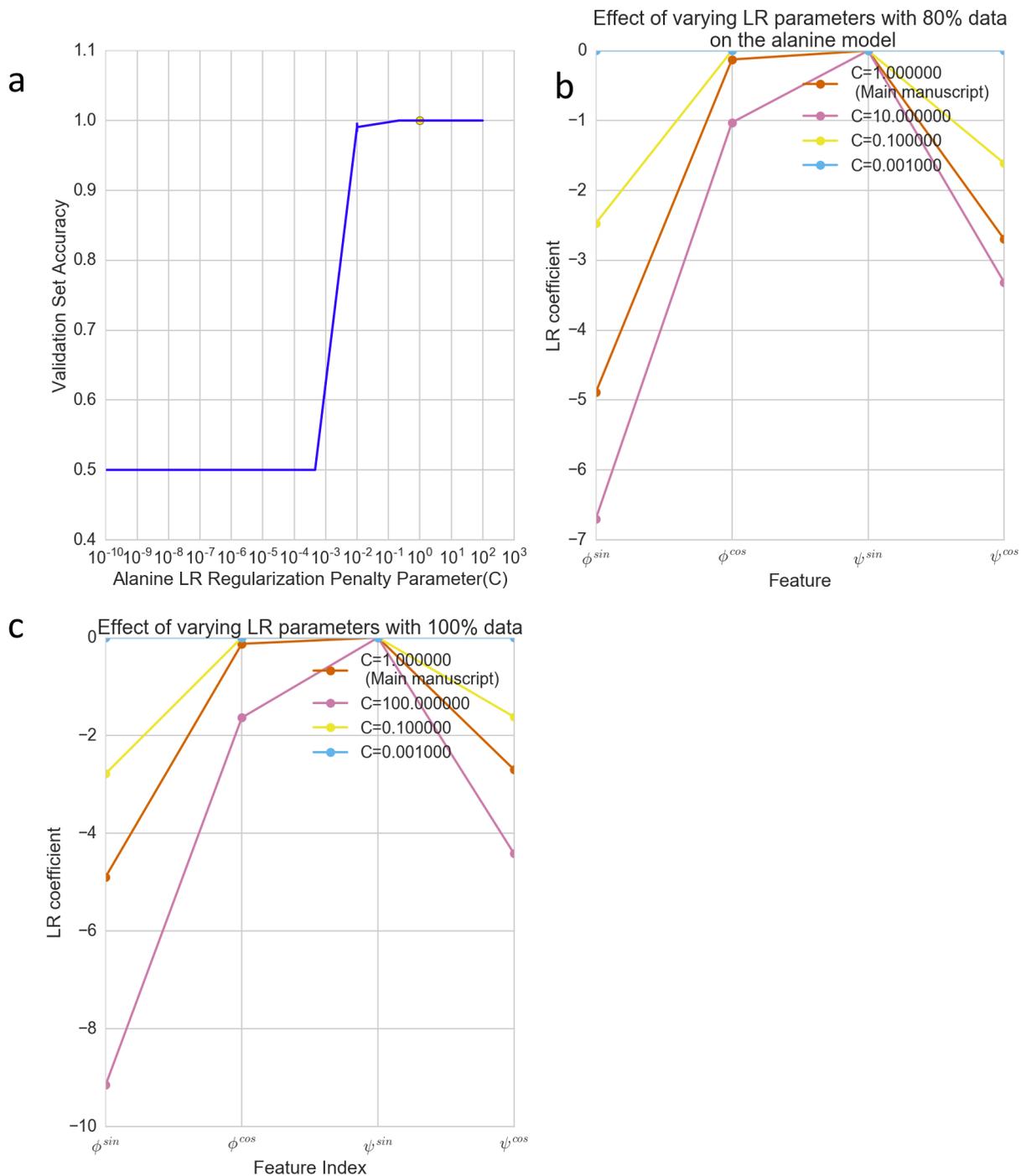

Figure 4: Sensitivity analysis for the L-1 regularized alanine LR model. a). Results from 3-fold cross-validation shows that the model has a very high validation accuracy (~1) across several magnitude changes in the C-parameter. The error bars are negligible in the validation dataset(n=3). The gold star is the model presented in the paper after it has been fit to the complete dataset. b). Changing the Logistic regressions's regularization/slack parameter by orders of magnitude has minimal effect on the LR coefficients. In this case, we randomly removed 20% of the data to show that the model was still robust to perturbations. c). This result remains consistent with 100% of the data across several magnitude changes in the slack parameter. It is worth noting that the LR's output is an inverse exponential transform of the input features which likely dampens the effects of these perturbations.

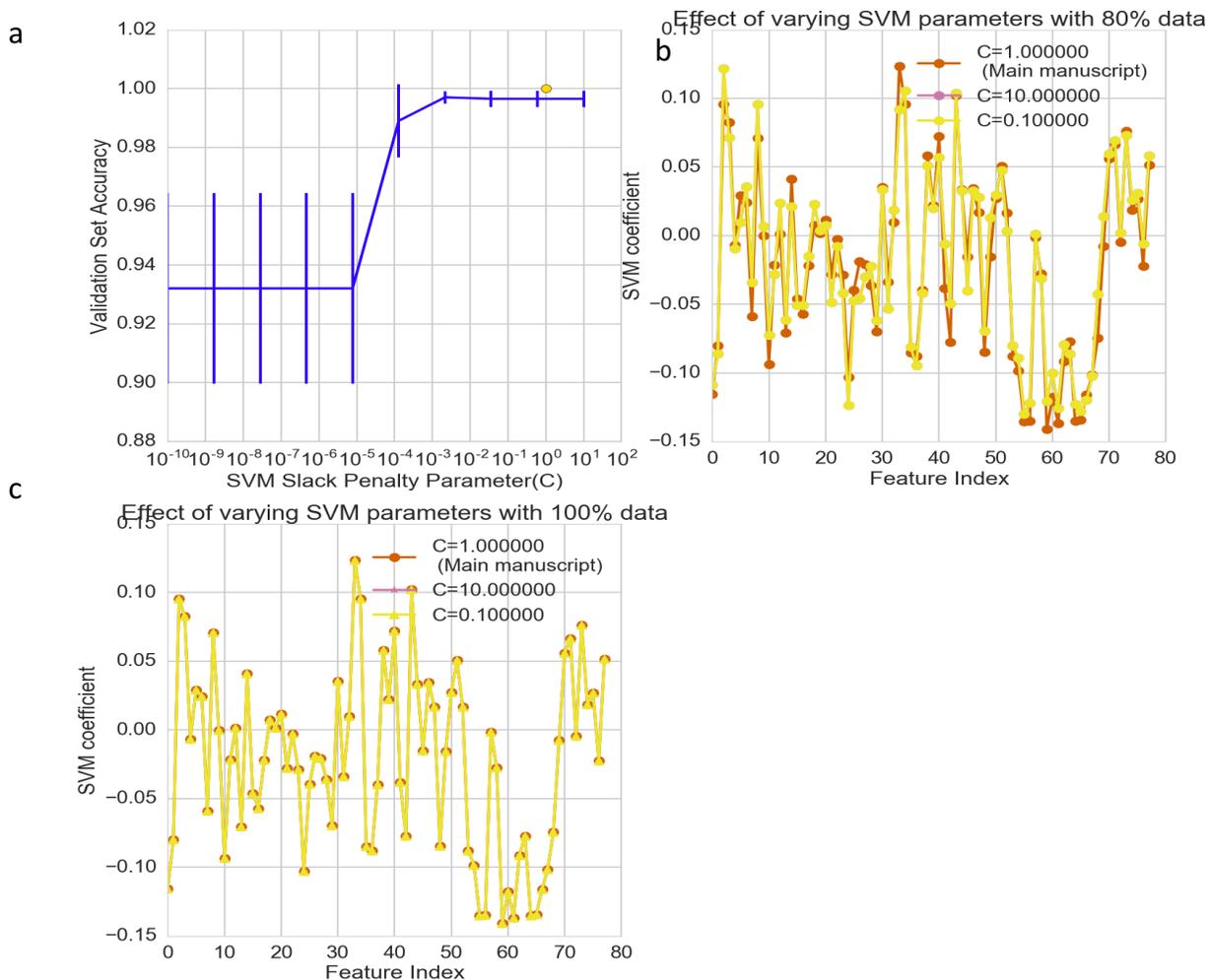

*Figure 5: Sensitivity analysis for the Chignolin model. a). Results from 3-fold cross-validation shows that the model has a very high validation accuracy (> .92) across several magnitude changes in the C-parameter. The error bars indicate standard deviation in the validation scores(n=3). The gold star is the model presented in the paper after it has been fit to the complete dataset. b). Changing the SVM's regularization/slack parameter C by an order of magnitude in either direction has minimal effect on the SVM coefficients. In this case, we also randomly removed 20% of the data to show that the model was robust to perturbations. c). This result remains consistent with 100% of the data across several magnitude changes in the C-parameter.*

**Supporting Note 1: Using Logistic Regression Classifiers to define CVs**

Similarly, the results from our $LR_{cv}$ simulations are shown in Figure 6. As it can be seen in Figure 6a-b, the LR model, similar to the SVM, learns a coordinate (Figure 6b) from just 4ns of training data. Running Metadynamics simulations along the predicted state probability, $LR_{cv}$, allows us to observe >10 transitions along alanine's slower $\phi$ coordinate in just 45ns of sampling. While the $LR_{cv}$ is less efficient than the $SMV_{cv}$, it is worth noting that in our previous study[14], only 2 transitions were observed in 170ns of unbiased MD simulations for solvated alanine dipeptide. Similar to the SVM model, the robust transition statistics combined with the reweighting method of Tiwary[11] now allow us to recover the free-energy landscape (Figure 6d) across other coordinates of interest.

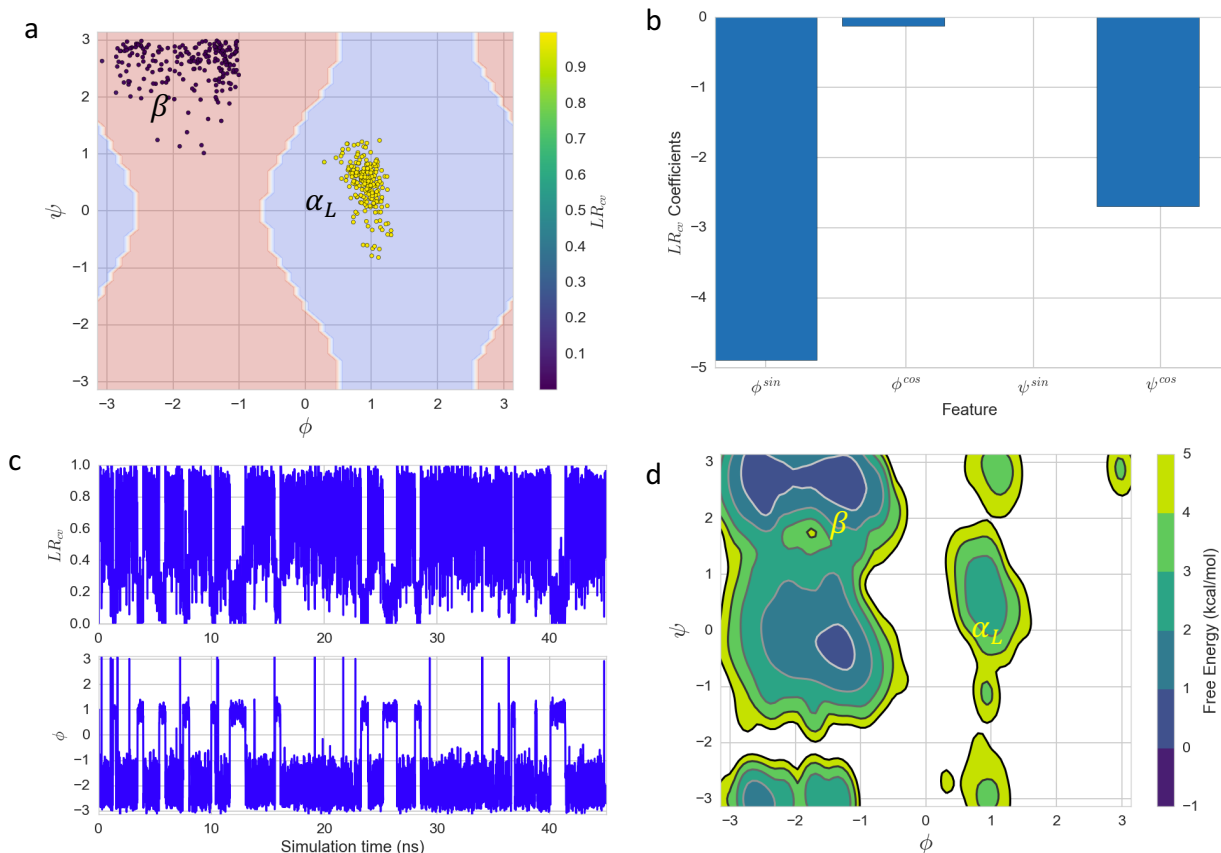

*Figure 6: Results from L-1 regularized $LR_{cv}$. a) The two 2ns training trajectories projected unto the Ramachandran plot. Note that no transition was observed from the $\beta$ to the $\alpha_L$ basins. The training frames are colored according to the assigned state probabilities as predicted via the trained logistic regression model. The contours show the decision boundaries for each region. b) Decomposing the logistic regressions' coefficients vector into the original features shows how the model learns non-trivial non-uniform mapping between the input features and output state labels. c) Running well-tempered Metadynamics simulations along the predicted state probabilities allows us efficiently sample the $\beta$ to $\alpha_L$ transition repeatedly. d. Reweighting allows us to recover the full free energy surface along the Ramachandran plot.*